\documentclass[
]{ceurart}

\usepackage{listings}
\usepackage{cleveref}
\usepackage{booktabs}
\usepackage{graphicx}
\usepackage{url}

\begin{document}

\copyrightyear{2026}
\copyrightclause{Copyright for this paper by its authors. Use permitted under Creative Commons License Attribution 4.0 International (CC BY 4.0).}

\conference{Second International TEXT2SPARQL Challenge, Co-Located with Text2KG at ESWC26, May 10, 2026, Dubrovnik, Croatia.}

\title{Towards Researcher Agents for Knowledge-Graph Question Answering}

\author[1]{Tommaso Soru}[email=tom@liberai.org]
\cormark[1]
\author[1]{Abdulsobur Oyewale}[email=abdulsobur@liberai.org]
\address[1]{Liber AI Research, London, United Kingdom}

\cortext[1]{Corresponding author.}

\begin{abstract}
Translating a natural-language question into a SPARQL query that can be executed against a large knowledge graph requires resolving lexical ambiguity, grounding surface terms in the target ontology, and producing graph patterns that are both syntactically valid and semantically faithful. We present an agentic text-to-SPARQL system that goes one step beyond static tool-using agents: a \emph{researcher agent} that, after each round of inference on a validation set, proposes and tests changes to its own prompts, rules, and tool-orchestration code. We instantiate the loop on DBpedia, evolve nine successive versions of the agent driven by a low-cost reasoning model, and deploy the best-performing configuration with two stronger backbone models. The study yields three observations: (i) self-improvement converges quickly and then achieves 0.22 overall accuracy on the 2025 DBpedia validation set; (ii) the bottleneck is consistently in basic-graph-pattern predicate selection, not in SPARQL syntax or modifiers; and (iii) several benchmark items appear to penalise correct queries due to property ambiguity in DBpedia, suggesting that future Text-to-SPARQL benchmarks should be scored using a combination of machine translation and information retrieval metrics.
\end{abstract}

\begin{keywords}
  Text-to-SPARQL \sep
  Knowledge Graph Question Answering \sep
  LLM Agents \sep
  Self-improving Agents \sep
  DBpedia
\end{keywords}

\maketitle

\section{Introduction}
\label{sec:intro}

Public knowledge graphs such as DBpedia~\cite{doi:10.3233/SW-140134}, Wikidata~\cite{wikidata_article}, and Freebase~\cite{bollacker2008freebase} expose billions of semantic triples and form one of the largest publicly available repositories of structured world knowledge. To make this knowledge accessible to non-expert users, question answering (QA) over knowledge graphs has long sought to replace formal query languages such as SPARQL with natural-language interfaces~\cite{SoruetAl:SEMPDS2017, soru2018neural, rony2022sgpt}. The Text-to-SPARQL task remains difficult: a system must disambiguate surface terms, ground them in a specific ontology, satisfy graph-structural constraints, and emit a syntactically valid query that often spans multiple hops, aggregations, and filters~\cite{cao-etal-2022-kqa, Liang2021-vm, dubey2019lc, 10.1145/3708326, app14041521}.

Early Text-to-SPARQL pipelines aligned parsed questions with fixed SPARQL skeletons, as in the Template-Based Question Answering over Linked Data (TBSL) framework~\cite{10.1145/2187836.2187923} and the keyword-driven template generator of~\cite{sparql_from_template}. Such systems require substantial manual effort and degrade quickly on paraphrases, unseen ontologies, and compositional queries. Semi-automatic methods reduce authoring effort by learning to rank or select templates from data, as in the skeleton-grammar approach with neural ranking proposed in SPARQA~\cite{Sun_Zhang_Cheng_Qu_2020}, but remain constrained by a finite template inventory. The First International TEXT2SPARQL Challenge~\cite{text2sparql2025} showed that LLM-based agents, often equipped with retrieval and KG-exploration tools, now outperform these template-based pipelines across heterogeneous KGs.

This paper extends our prior submission to the First TEXT2SPARQL Challenge~\cite{ref:Soru2025}, which fine-tuned autoregressive code models on a large synthetic corpus, by changing the level at which learning happens. Rather than fine-tuning model weights, we let the agent edit its own source code between evaluation rounds. Concretely, our contributions are: (i) a \emph{researcher agent} architecture in which an LLM proposes changes to the prompts, rules, and tool orchestration of a downstream Text-to-SPARQL agent and tests them against a validation set; (ii) a pre-evaluation that compares six recent LLMs (open and closed) on the 2025 DBpedia validation set and motivates the choice of one cheap model for self-improvement and two stronger models for the final challenge submission; and (iii) a feature-construction study over nine successive agent versions that isolates which rules and tool-grounding heuristics actually move accuracy. The agent and traces are released as open artefacts at \url{https://github.com/LiberAI/researcher-agents-for-kgqa}.

The remainder of the paper is structured as follows. \Cref{sec:relatedwork} reviews related work. \Cref{sec:proposal} describes the static and self-improving agent loops and the researcher agent we built. \Cref{sec:eval} reports the pre-evaluation, the feature-construction study, and an inspection of the benchmark itself. \Cref{sec:conclusion} concludes.

\section{Related Work}
\label{sec:relatedwork}

\paragraph{Neural Text-to-SPARQL.}
A large body of recent work casts Text-to-SPARQL as a neural sequence-generation problem. Pre-trained sequence-to-sequence models such as T5 and BART arrange entities and relations into well-formed SPARQL queries with little task-specific machinery~\cite{banerjee2022modern}. SGPT~\cite{rony2022sgpt} stacks Transformer encoders on top of a GPT-2 decoder but tends to mis-handle multi-hop patterns. Prompt-based approaches that inject relevant subgraphs into the model's context, such as SPARQLGEN~\cite{kovriguina2023sparqlgen}, still exhibit systematic ordering errors. A triplet-order-sensitive pre-training scheme for T5 was proposed in~\cite{app14041521} to alleviate this issue. These methods generally rely on costly fine-tuning and specialised decoding strategies, which makes them expensive to port across knowledge graphs~\cite{kovriguina2023sparqlgen, app14041521}.

\paragraph{End-to-end systems over DBpedia.}
Several end-to-end models have been evaluated on DBpedia-centric benchmarks. SPARKLE~\cite{lee2025sparkle} integrates KG structure into the decoder and reports state-of-the-art F1 on LC-QuAD~1.0. A hybrid multi-head convolutional encoder combining CNNs and self-attention~\cite{chen2024integrating} obtains strong BLEU and F1 on QALD-9 and LC-QuAD~1.0. A universal KG-QA model that frames question understanding as sequence generation~\cite{omar2023universal} is competitive on LC-QuAD~1.0 and QALD-9. A two-stage ontology-guided prompting framework that first predicts a SPARQL skeleton and then fills in KG-specific information reaches 79.1\% F1 on LC-QuAD~1.0~\cite{jiang2025ontology}. Earlier work introduced a silhouette-based two-step architecture that combines coarse query generation with graph search~\cite{purkayastha2022deep}.

\paragraph{Benchmarks and synthetic datasets.}
Large-scale QA benchmarks have driven much of this progress. LC-QuAD~2.0 contains roughly 30k complex natural-language questions paired with SPARQL queries over DBpedia and Wikidata~\cite{dubey2019lc}, while the QALD series provides smaller, multilingual benchmarks; QALD-9-Plus extends QALD-9 with high-quality DBpedia translations into eight languages and five under-represented languages including Ukrainian, Armenian, Lithuanian, Bashkir, and Belarusian~\cite{perevalov2022qald}. The DBpedia Neural QA (DBNQA) corpus offers approximately 894{,}499 English question--SPARQL pairs generated via templating~\cite{hartmann2018generating}.

\paragraph{First TEXT2SPARQL Challenge.}
Submissions to the First TEXT2SPARQL Challenge~\cite{text2sparql2025} explored a wide spectrum of strategies. Wardenga and K\"afer~\cite{ref:Wardenga2025} exploit data shapes in the LLM context to constrain generation on public and private KGs. Our prior submission~\cite{ref:Soru2025} fine-tunes autoregressive code models on a large synthetic corpus. Dorsch et al.~\cite{ref:Dorsch2025} build an organisational KGQA agent. ARUQULA~\cite{ref:Brei2025} combines an LLM with ReAct-style reasoning~\cite{yao2023react} and KG-exploration utilities. The AIRI team~\cite{ref:Berezin2025} proposes a Text-to-SPARQL executor that decouples generation from execution. Perevalov and Both~\cite{ref:Perevalov2025} push the multilingual frontier with human-inspired reasoning.

\paragraph{Agentic and self-improving LLM systems.}
Beyond Text-to-SPARQL, recent work on agentic LLM systems shows that interleaving reasoning with tool use improves grounding and reduces hallucinations. ReAct~\cite{yao2023react} demonstrates this synergy on QA and decision-making benchmarks, and Toolformer~\cite{schick2023toolformer} shows that LLMs can be self-supervised to call external APIs. Schluntz and Zhang's practitioner-oriented overview~\cite{schluntz2024building} distinguishes between workflows and agents along the dimensions of inference loops and tool use. Self-improvement adds a second loop, in which the agent edits its own scaffolding rather than only its scratchpad; our researcher agent is an instance of this pattern, specialised to Text-to-SPARQL.


\section{Proposal}
\label{sec:proposal}

\subsection{From Static to Self-Improving Agents}

A static Text-to-SPARQL agent realises a single \emph{inference loop}: given a user request, the LLM emits an action (a tool call or a SPARQL draft), the environment (lookup indices, the SPARQL endpoint) returns feedback, and the LLM either issues a further action or returns a final query. The scaffolding---prompts, rules, tool wiring---is fixed across queries.

A self-improving agent adds a second, outer loop. After each round of inference on a validation set, an orchestrator LLM observes the agent's failures and successes and \emph{proposes a change to the agent's source code}. The change is applied to a working copy, the new agent is tested on the same validation set, and the loop continues with the human gradually moved out of the design path. We use the term \emph{researcher agent} to emphasise that the orchestrator's role is closer to that of an ML researcher running a feature-construction study than to a chain-of-thought reasoner.

\subsection{Researcher Agent for Text-to-SPARQL}

The downstream Text-to-SPARQL agent consumes a natural-language question and a target SPARQL endpoint and exposes three tools to the LLM: (i) a \emph{surface-form lookup} that maps spans of the question to candidate DBpedia entities, (ii) an \emph{ontology lookup} that enumerates classes and predicates around a candidate entity, and (iii) a \emph{test-SPARQL} tool that executes a draft query against the endpoint and returns either the answer or the error message. The agent iterates plan/ground/draft/verify steps until either a query returns a non-empty result that matches the expected answer shape or an iteration budget is exhausted.

The researcher agent wraps this Text-to-SPARQL agent and, between evaluation rounds, edits its prompts, its small set of hard rules (e.g.,\ ``prefer \texttt{dbo:} over \texttt{dbp:}''), the number of in-context examples, and small pieces of the tool-orchestration code. Each edit produces a new agent version $v_i$; the researcher records every version, its diff, and its scores. Across our study the agent evolved nine times, alternately removing and reinstating rules according to whether they improved overall accuracy or only specific SPARQL sub-components.

\section{Experimental Evaluation}
\label{sec:eval}

\subsection{Datasets}

We use the DBpedia validation set from the First International TEXT2SPARQL Challenge (henceforth DB25) for both the LLM pre-evaluation and the feature-construction study, and we submit the best agent configuration to the Second International TEXT2SPARQL Challenge with two stronger backbones.

\subsection{Experimental Settings}

The Text-to-SPARQL agent is implemented in Python on top of an open-source agent framework. The researcher agent uses the same framework. All tool calls during evaluation are issued against the official challenge SPARQL endpoint. For each agent version we report the official challenge metrics broken down into four families: Basic Graph Pattern (BGP) nodes (entity grounding), BGP predicates (relation grounding), inner operators (modifiers such as \texttt{DISTINCT} or \texttt{COUNT}) and outer operators (overall query structure).

\subsection{LLM Pre-evaluation}

We first ran the initial agent (v1) on DB25 under six recent LLMs spanning the cost/capability frontier: Gemini~3 Flash, Claude Sonnet 4.6, GPT-5.4 Mini, DeepSeek~v3.2, Qwen~3.5 122B, and Gemma~3 27B. Overall accuracies clustered between 0.16 and 0.23, with Qwen~3.5 122B leading the open-weights bracket and Claude Sonnet 4.6 leading the closed bracket. We selected DeepSeek~v3.2 as the cheapest reasoning-enabled model for the feature-construction study, and used Claude Sonnet 4.6 and Qwen 3.5 122B for the final challenge submission.

\subsection{Feature-Construction Study}

\Cref{tab:features} summarises the nine agent versions produced by the researcher agent on DeepSeek~v3.2. Each column corresponds to a version; each row to a binary feature toggled by the researcher. The bottom row reports overall accuracy on DB25.

\begin{table}[t]
\centering
\caption{Feature-construction study on the DB25 validation set using DeepSeek~v3.2 with reasoning enabled.}
\label{tab:features}
\footnotesize
\begin{tabular}{lccccccccc}
\toprule
Characteristic & v1 & v2 & v3 & v4 & v5 & v6 & v7 & v8 & v9 \\
\midrule
\texttt{dbo:} preference rule        & \checkmark & -- & \checkmark & -- & \checkmark & \checkmark & \checkmark & \checkmark & \checkmark \\
\texttt{SELECT DISTINCT}             & -- & -- & \checkmark & \checkmark & \checkmark & \checkmark & \checkmark & \checkmark & \checkmark \\
Grouped \texttt{dbo}/\texttt{dbp} pairs in context & -- & -- & -- & -- & \checkmark & \checkmark & \checkmark & \checkmark & \checkmark \\
Triple counts shown in context       & -- & -- & -- & \checkmark & -- & \checkmark & \checkmark & \checkmark & \checkmark \\
Triple counts used for selection     & -- & -- & -- & \checkmark & -- & -- & -- & -- & -- \\
Unicode entity-linking fallback      & -- & -- & -- & -- & -- & \checkmark & \checkmark & \checkmark & \checkmark \\
\texttt{rdf:type} constraints rule   & -- & -- & -- & -- & -- & \checkmark & \checkmark & \checkmark & \checkmark \\
Multi-hop \texttt{rdf:type} exception & -- & -- & -- & -- & -- & -- & -- & -- & \checkmark \\
\texttt{COUNT} format (no \texttt{AS} alias) & -- & -- & -- & -- & -- & \checkmark & \checkmark & \checkmark & \checkmark \\
Triple direction rule                & -- & -- & -- & -- & -- & -- & -- & -- & \checkmark \\
Numbered / structured rules          & -- & -- & -- & -- & -- & -- & \checkmark & -- & -- \\
No.\ of examples (N-shot)            & 3 & 3 & 3 & 3 & 5 & 5 & 7 & 6 & 6 \\
\midrule
\textbf{Overall accuracy}            & 0.20 & 0.06 & 0.18 & 0.12 & 0.18 & \textbf{0.22} & 0.17 & \textbf{0.22} & 0.21 \\
\bottomrule
\end{tabular}
\end{table}

\Cref{tab:breakdown} reports the per-section breakdown. Inner-operator accuracy is already high in v1 (0.84) and stays in the 0.79--0.86 band throughout, indicating that LLMs handle SPARQL modifiers reliably. Outer-operator accuracy rises from 0.75 to 0.86 as \texttt{SELECT DISTINCT}, the \texttt{COUNT} format fix, and the \texttt{rdf:type} constraints rule are introduced. BGP node accuracy is the most volatile family, jumping from 0.25 in v1 to 0.67 in v4 once triple-count cues are added to the context, then settling around 0.59 once those cues are kept in the context but no longer used directly for selection. BGP predicates remain the bottleneck across the entire study, oscillating between 0.20 and 0.31. The overall metric achieves 0.22 from v6 onwards and the researcher agent's edits stop translating into gains.

\begin{table}[t]
\centering
\caption{Per-section accuracy breakdown on DB25 using DeepSeek~v3.2.}
\label{tab:breakdown}
\footnotesize
\begin{tabular}{lccccc}
\toprule
Agent & BGP Nodes & BGP Predicates & Inner ops & Outer ops & Overall \\
\midrule
v1 & 0.25 & 0.25 & 0.84 & 0.75 & 0.20 \\
v2 & 0.57 & 0.26 & 0.81 & 0.52 & 0.06 \\
v3 & 0.61 & 0.26 & 0.79 & 0.76 & 0.18 \\
v4 & \textbf{0.67} & 0.20 & 0.85 & 0.74 & 0.12 \\
v5 & 0.55 & 0.28 & 0.85 & 0.77 & 0.18 \\
v6 & 0.59 & 0.29 & \textbf{0.86} & \textbf{0.86} & \textbf{0.22} \\
v7 & 0.61 & 0.22 & 0.85 & 0.84 & 0.17 \\
v8 & 0.59 & \textbf{0.31} & 0.82 & 0.81 & \textbf{0.22} \\
v9 & 0.59 & 0.27 & 0.85 & 0.84 & 0.21 \\
\bottomrule
\end{tabular}
\end{table}

We picked v6 as the agent configuration to deploy for the Second TEXT2SPARQL Challenge, on the grounds that it ties for the highest overall accuracy while also leading on both operator families, and that v8 simply reverts v7 onto v6. The gap in performance metrics is due to non-determinism in LLMs.

\subsection{Inspecting the Benchmark}
\label{sec:benchmark}

The result of 0.22 is well below what one would expect from a modern reasoning LLM with grounded tools, and it raised a natural question: are the remaining errors really agent failures, or are they benchmark failures? A manual inspection of the questions that v6 missed pointed to a recurring pattern of property ambiguity in DBpedia. As an example, the question \emph{``How many unique authors have written science fiction novels?''} carries the expected query
\begin{lstlisting}
SELECT DISTINCT COUNT(?author) WHERE {
  ?x dbo:literaryGenre dbr:Science_fiction .
  ?x dbo:author ?author .
}
\end{lstlisting}
which returns 2{,}892 results. The Claude Sonnet 4.6 agent instead produced
\begin{lstlisting}
SELECT COUNT(DISTINCT(?author) AS ?c) WHERE {
  ?x dbp:genre dbr:Science_fiction .
  ?x dbo:author ?author .
}
\end{lstlisting}
returning 1{,}116 results. Both queries are arguably correct natural-language renderings: \texttt{dbo:literaryGenre} and \texttt{dbp:genre} co-exist in DBpedia for overlapping but non-identical sets of works, and ordering \texttt{DISTINCT} before \texttt{COUNT} is itself ambiguous. A second attempt on the same item added an \texttt{rdf:type dbo:Novel} constraint and returned 0 results because few works are typed as novels in DBpedia.

These observations motivate treating Text-to-SPARQL evaluation as a machine-translation task with structural credit assignment, rather than as information retrieval over the endpoint---a view we previously advocated in \emph{SPARQL as a Foreign Language}~\cite{SoruetAl:SEMPDS2017} and which appears even more pressing for agentic systems whose tool feedback can drive them away from the reference query for the right reason. A combination of machine translation and information retrieval metrics may be optimal.

\subsection{Results and Discussion}

\begin{table}[t]
\centering
\caption{Accuracy of the initial agent across LLM backbones on DB25.}
\label{tab:model-accuracy}
\footnotesize
\begin{tabular}{lccccc}
\toprule
Model & Match & Nodes & Preds & Inner & Outer \\
\midrule
Gemini 3 Flash Preview & 0.21 & 0.26 & 0.26 & 0.79 & \textbf{0.79} \\
Claude Sonnet 4.6 & 0.21 & 0.27 & 0.27 & 0.66 & 0.75 \\
GPT-5.4 Mini & 0.20 & 0.24 & 0.24 & \textbf{0.84} & 0.77 \\
DeepSeek~v3.2 & 0.20 & 0.25 & 0.25 & \textbf{0.84} & 0.75 \\
Qwen 3.5 122B & \textbf{0.23} & \textbf{0.29} & \textbf{0.29} & 0.79 & 0.73 \\
Gemma 3 27B & 0.15 & 0.17 & 0.17 & 0.82 & 0.74 \\
\bottomrule
\end{tabular}
\end{table}

\begin{figure}[t]
\centering
\includegraphics[width=0.8\linewidth]{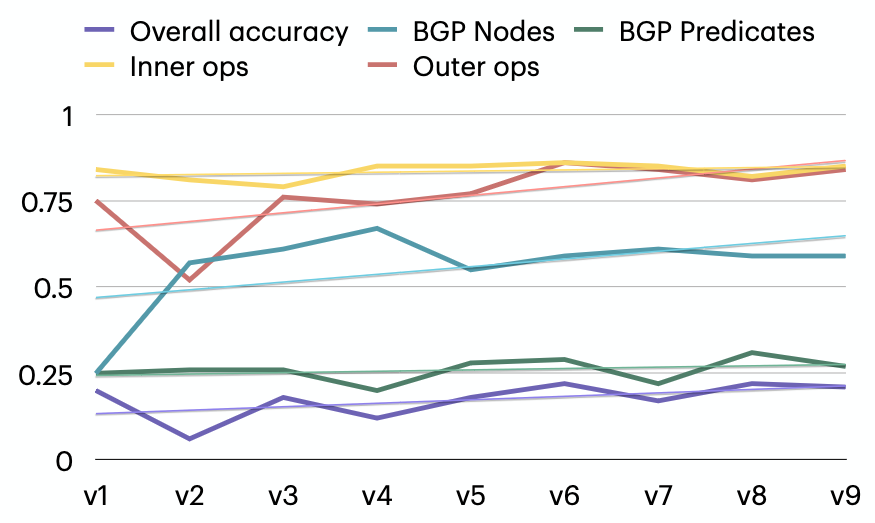}
\caption{Overall DB25 accuracy across the nine DeepSeek~v3.2 researcher-agent versions.}
\label{fig:v1-v9-accuracy}
\end{figure}

\begin{table}[t]
\centering
\caption{Final evaluation results on CK26 and DB26, as obtained by the challenge organisers.}
\label{tab:final-evaluation}
\footnotesize
\resizebox{\linewidth}{!}{%
\begin{tabular}{llccccc}
\toprule
Dataset & System & F1-score/NDCG & Recall & Precision & F1-score & NDCG \\
\midrule
CK26 English & LIBER-AI-CLAUDE & 0.320820 & 0.366165 & 0.315080 & 0.321504 & 0.348168 \\
CK26 English & LIBER-AI-QWEN   & 0.174532 & 0.207819 & 0.166049 & 0.172381 & 0.331639 \\
\midrule
CK26 German  & LIBER-AI-CLAUDE & 0.349699 & 0.419990 & 0.336803 & 0.350960 & 0.348168 \\
CK26 German  & LIBER-AI-QWEN   & 0.239887 & 0.255296 & 0.238459 & 0.238952 & 0.348168 \\
\midrule
DB26 English & LIBER-AI-CLAUDE & 0.323479 & 0.335960 & 0.336032 & 0.323479 & -- \\
DB26 English & LIBER-AI-QWEN   & 0.312932 & 0.319897 & 0.311046 & 0.312932 & -- \\
\midrule
DB26 Spanish & LIBER-AI-CLAUDE & 0.373026 & 0.394291 & 0.365254 & 0.373026 & -- \\
DB26 Spanish & LIBER-AI-QWEN   & 0.236129 & 0.239048 & 0.238528 & 0.236129 & -- \\
\bottomrule
\end{tabular}%
}
\end{table}

\Cref{tab:model-accuracy} shows that changing the LLM backbone produces only modest variation in exact match: the best model, Qwen 3.5 122B, reaches 0.23, while the remaining competitive systems cluster around 0.20--0.21. The same table also confirms the error profile observed in the feature-construction study: inner and outer operators are usually much easier than BGP grounding, whereas node and predicate scores remain below 0.30 for all models. \Cref{fig:v1-v9-accuracy} makes the self-improvement dynamics explicit. Accuracy is non-monotonic across agent versions, with harmful edits in v2 and v4, recovery by v6, but the positive trendlines suggest that the accuracy could have improved further beyond 0.22. \Cref{tab:final-evaluation} reports the final CK26 and DB26 results available so far: Claude outperforms Qwen on both datasets and all reported languages, with the strongest combined F1-score/NDCG observed on DB26 Spanish. Taken together, the model comparison, the version trajectory, and the final evaluation suggest that the main remaining gains are unlikely to come from prompt-level edits alone; they require better grounding of DBpedia entities and, especially, predicates. The reason we achieve a higher F1-score with these two models in DB26 (0.34, 0.31) than in DB25 (0.21, 0.23) is likely due to the DB26 annotators solving several issues in the data.

\section{Conclusion and Future Work}
\label{sec:conclusion}

We presented a researcher-agent approach to Text-to-SPARQL in which an orchestrator LLM iteratively edits the source code of a downstream Text-to-SPARQL agent on a validation set. The study converged after nine versions to a configuration that ties for the best overall accuracy on the 2025 DBpedia validation set while leading both operator families. The low accuracy at 0.22 appears to be driven, at least in part, by ambiguities and errors in the benchmark itself rather than by agent failures: many of the remaining misses correspond to plausible alternative renderings of the question in DBpedia's ontology. However, positive trendlines suggest that the accuracy could have improved further beyond that value. Future work includes (i) extending the researcher agent to private KGs, (ii) learning the orchestrator's editing policy from execution traces rather than hand-crafted meta-prompts, and (iii) collaborating with the community on an updated Text2SPARQL benchmark scored at the structural level rather than purely by endpoint results.

\section*{Acknowledgments}

The authors thank the organisers of the Second TEXT2SPARQL Challenge and the contributors who raised methodology issues on the 2025 DBpedia dataset.

\section*{Declaration on Generative AI}

The authors used generative AI tools to assist with phrasing and editing portions of this manuscript. All technical content, experiments, and conclusions were verified and approved by the authors, who take full responsibility for the publication.

\bibliography{paper}

\end{document}